\documentclass{article} 
\usepackage{nips13submit_e,times}
\usepackage{hyperref}
\usepackage{url}
\usepackage{graphicx}
\usepackage{lineno,epstopdf,amsmath,color,appendix}

\title{Reservoir Computing using Cellular Automata}

\author{
Ozgur ~Yilmaz\thanks{ ozguryilmazresearch.net} \\
Department of Computer Engineering\\
Turgut Ozal University\\
Ankara, Turkey \\
\texttt{ozyilmaz@turgutozal.edu.tr} \\
}

\nipsfinalcopy 

\begin{document}

\maketitle

\begin{abstract}
We introduce a novel framework of reservoir computing. Cellular automaton is used as the reservoir of dynamical systems. Input is randomly projected onto the initial conditions of automaton cells and nonlinear computation is performed on the input via application of a rule in the automaton for a period of time. The evolution of the automaton creates a space-time volume of the automaton state space, and it is used as the reservoir. The proposed framework is capable of long short-term memory and it requires orders of magnitude less computation compared to Echo State Networks. Also, for additive cellular automaton rules, reservoir features can be combined using Boolean operations, which provides a direct way for concept building and symbolic processing, and it is much more efficient compared to state-of-the-art approaches.

\end{abstract}

\section{Introduction}

Many real life problems in artificial intelligence require the system to remember previous inputs. Recurrent neural networks (RNN) are powerful tools of machine learning with memory. For this reason they have become one of the first choices for modeling dynamical systems. In this paper we propose a novel recurrent computation framework that is analogous to Echo State Networks (ESN) but with very low computational complexity. The proposed algorithm uses cellular automata in Reservoir Computing (RC) architecture and is capable of Long-Short-Term-Memory (LSTM). Additionally, the binary nature of the feature space and additivity of the cellular automaton rules enable Boolean logic, and provides a great potential for symbolic processing. In the following sections we review reservoir computing, cellular automata and neuro-symbolic computation, then we state the contribution of our study.  

\subsection{Reservoir Computing}

Recurrent Neural Networks are connectionist computational models that utilize distributed representation and nonlinear dynamics of its units. Information in RNNs is propagated and processed in time through the states of its hidden units, which make them appropriate tools for sequential information processing. 

RNNs are known to be Turing complete computational models \cite{siegelmann1995computational} and universal approximators of dynamical systems \cite{funahashi1993approximation}. They are especially appealing for problems that require remembering long-range statistical relationships such as speech, natural language processing, video processing, financial data analysis etc. 

Despite their immense potential as universal computers, difficulties in training RNNs arise due to the inherent difficulty of learning long-term dependencies \cite{hochreiter1991untersuchungen,bengio1994learning,hochreiter1997long} and convergence issues \cite{doya1992bifurcations}. However, recent advances suggest promising approaches in overcoming these issues, such as utilizing a reservoir of coupled oscillators \cite{jaeger2001echo,maass2002real}.

Reservoir computing (echo state networks or liquid state machines) alleviates the problem of training in a recurrent network by using a static dynamical reservoir of coupled oscillators, which are operating at the edge of chaos. It is claimed that many of these type of dynamical systems possess high computational power \cite{bertschinger2004real,legenstein2007edge}. In this approach, due to rich dynamics already provided by the reservoir, there is no need to train many recurrent layers and learning takes place only at the output (or read-out stage) layer. This simplification enables usage of recurrent neural networks in complicated tasks that require memory for long-range (both spatially and temporally) statistical relationships.  

The essential feature of the network in the reservoir is called echo state property \cite{jaeger2001echo}. In networks with this property, the effect of previous state and previous input dissipates gradually in the network without getting amplified. In classical echo state networks, the network is generated randomly and sparsely, considering the spectral radius requirements of the weight matrix. Even though spectral radius constraint ensures stability of the network to some extent, it does not say anything about the short-term memory or computational capacity of the network. The knowledge about this capacity is essential for proper design of the reservoir for the given task. 

The reservoir is expected to operate at the edge of chaos because the dynamical systems are shown to present high computational power at this mode \cite{bertschinger2004real,legenstein2007edge}. High memory capacity is also shown for reservoirs at the edge of chaos. Lyapunov exponent is a measure edge of chaos operation in a dynamical system, and it can be empirically computed for a reservoir network \cite{legenstein2007edge}.  However, this computation is not trivial or automatic, and needs expert intervention \cite{lukovsevivcius2009reservoir}. 

It is empirically shown that there is an optimum Lyapunov exponent of the reservoir network, related to the amount of memory needed for the task \cite{verstraeten2007experimental}. Thus, fine-tuning the connections in the reservoir for learning the optimal connections that lead to optimal Lyapunov exponent is very crucial for achieving good performance with the reservoir. There are many types of learning methods proposed for tuning the reservoir connections (see \cite{lukovsevivcius2009reservoir} for a review), however optimization procedure on the weight matrix is prone to get stuck at local optimum due to high curvature in the weight space. 
 
The input in a complex task is generated by multiple different processes, for which the dynamics and spatio-temporal correlations might be very different. One important shortcoming of the classical reservoir computing approach is its inability to deal with multiple spatio-temporal scales simultaneously. Modular reservoirs have been proposed that contain many decoupled sub-reservoirs operating in different scales, however fine tuning the sub-reservoirs according to the task is not trivial. 

\subsection{Cellular Automata}

Cellular automaton is a discrete computational model consisting of a regular grid of cells, each in one of a finite number of states. The state of an individual cell evolves in time according to a fixed rule, depending on the current state and the state of neighbors. The information presented as the initial states of a grid of cells is processed in the state transitions of cellular automaton and computation is extremely local. Cellular automata governed by some of the rules are proven to be computationally universal, capable of simulating a Turing machine \cite{cook2004universality}. 

The rules of cellular automata are classified \cite{wolfram2002new} according to their behavior: attractor, oscillating, chaotic, and edge of chaos. Some of the rules in the last class are shown to be Turing complete (rule 110, Conway’s game of life). Lyapunov exponent of a cellular automaton can be computed and it is shown to be a good indicator of computational power of the automata \cite{baetens2010phenomenological}. A spectrum of Lyapunov exponent values can be achieved using different cellular automata rules. Therefore a dynamical system with specific memory capacity (i.e. Lyapunov exponent value) can be constructed by using a corresponding cellular automaton.  

Cellular automata have been previously used for associative memory and classification tasks. Tzionas et al. \cite{tzionas1994new} proposed a cellular automaton based classification algorithm. Their algorithm clusters 2D data using cellular automata, creating boundaries between different seeds in the 2D lattice. The partitioned 2D space creates geometrical structure resembling a Voronoi diagram. Different data points belonging to the same class fall into the same island in the Voronoi structure, hence are attracted to the same basin. Clustering property of cellular automata is exploited in a family of approaches, using rules that form attractors in lattice space \cite{chady1998evolution,ganguly2003survey}. The attractor dynamics of cellular automata resembles Hopfield network architectures \cite{hopfield1982neural}. These approaches have two major problems: low dimensionality and low computational power. The first problem is due to the need for representing data in 2D space and the need for non-trivial algorithms in higher dimensions. The second problem is due to limiting the computational representation of cellular automata activity with attractor dynamics and clustering. The time evolution of the cellular automata has very high computational representation, especially for edge of chaos dynamics, but this is not exploited if the presented data are classified according to the converged basin in 2D space. 

Another approach is cellular neural networks \cite{chua1988cellular}. It has been shown that every binary cellular automata  of any dimension is a special case of a cellular neural network of the same neighborhood size \cite{chua2002nonlinear}. However, cellular neural networks impose a very specific spatial structure and they are generally implemented on specialized hardware, generally for image processing (see \cite{javier2013efficient} for a recent design).

\subsection{Symbolic Processing on Neural Representations}
Uniting the expressive power of mathematical logic and pattern recognition capability of neural networks has been an open question for decades, although several successful theories have been proposed \cite{pollack1990recursive,van2006neural,bader2008connectionist}. Difficulty arises due to the very different mathematical nature of logical reasoning and dynamical systems theory. Recently Jaeger proposed a novel framework called "Conceptors" based on reservoir computing architecture \cite{jaeger2014controlling}. The Conceptors are linear operators learned from the activities of the reservoir neurons and they can be combined by elementary logical operators, which enables them to form symbolic representations of the neural activities and build semantic hierarchies. In a similar flavor, Mikolov et al. \cite{mikolov2013distributed} successfully used neural network representations of words (language modeling) for analogical reasoning. 

\subsection{Contributions}
We provide a very low computational complexity method for implementing reservoir computing based recurrent computation, using cellular automata. Cellular automata replace the echo state neural networks. This approach provides both theortical and practical advantages over classical neuron-based reservoir computing. We show that the proposed framework is capable of accomplishing long-short-term-memory tasks such as the famous 5 bit and 20 bit memory, which are known to be problematic for feedforward architectures \cite{hochreiter1997long}. Additionally we show that the framework has great potential for symbolic processing such that the cellular automata feature space can directly be combined by Boolean operations, hence they can represent concepts and form a hierarchy of semantic interpretations. The computational complexity of the framework is shown to be orders of magnitude lower than echo state network based reservoir computing approaches. In the next section we give the details of the algorithm and then provide results on pathological learning tasks.

\section{Methods}
\label{method_section}
In our reservoir computing method, data are passed on a cellular automaton instead of an echo state network and the nonlinear dynamics of cellular automaton provide the necessary projection of the input data onto an expressive and discriminative space.  Compared to classical neuron-based reservoir computing, the reservoir design is trivial: cellular automaton rule selection. Utilization of ‘edge of chaos’ automaton rules ensures Turing complete computation in the reservoir, which is not guaranteed in classical reservoir computing approaches.

\begin{figure*}
\begin{center}
\includegraphics[width=0.8\textwidth]{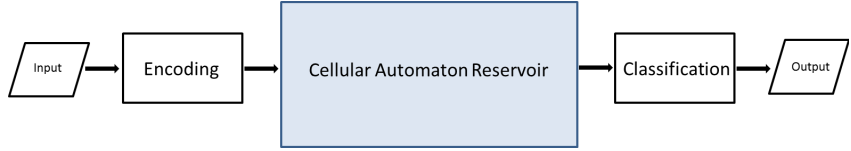}
\end{center}
   \caption{General framework for cellular automata based reservoir computing.}
\label{fig:fig1}
\end{figure*}

Algorithmic flow of our method is shown in Figure \ref{fig:fig1}. The reservoir computing system receives the input data. The encoding stage translates the input into the initial states of a 1D or multidimensional cellular automaton (2D is shown as an example). In reservoir computing stage, the cellular automaton rules are executed for a fixed period of iterations ($I$), to evolve the initial states. The evolution of the cellular automaton is recorded such that, at each time step a snapshot of the whole states in the cellular automaton is vectorized and concatenated. This output is a projection of the input onto a nonlinear cellular automata state space. Then the  cellular automaton output is used for further processing according to the task (eg. classification, compression, clustering etc.).

\begin{figure*}
\begin{center}
\fbox{\includegraphics[width=0.8\textwidth]{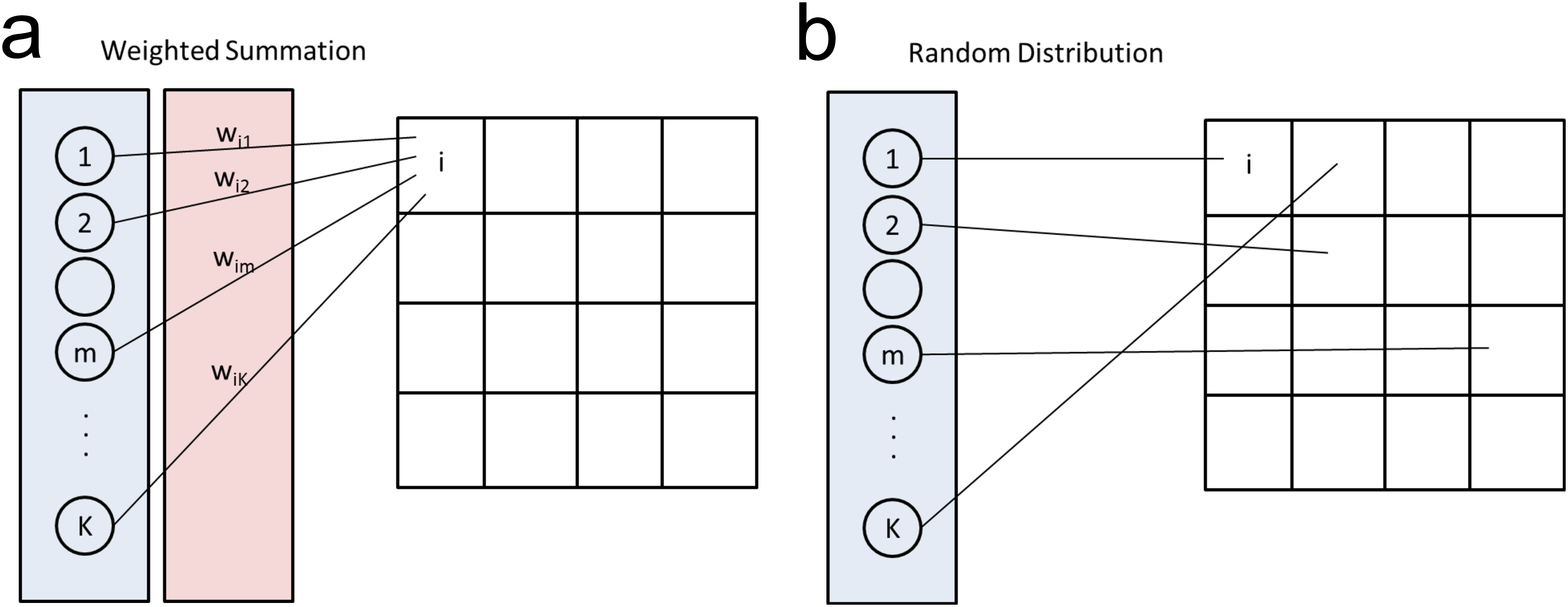}} 
\end{center}
   \caption{Two options for encoding the input into cellular automaton initial states. \textbf{a}. Each cell receives a weighted sum of the input dimensions. \textbf{b}. Each feature dimension is randomly mapped onto the cellular automaton cells. }
\label{fig:fig2}
\end{figure*}

In encoding stage there are two proposed options depending on the input data. 
\textbf{1.} For non-binary input data, each cell of cellular automaton might receive weighted input from every feature dimension of the input (Figure \ref{fig:fig2}a). The weighted sum is then binarized for each cell. In this option, instead of receiving input from the whole set of feature dimensions, a single cell can receive input from a subset of feature dimensions. In that case, the weight vector for a cell is sparse and a subspace of the input is processed by specific cells. In general, the weights can be set randomly as in echo state networks. 
\textbf{2.} For binary input data, each feature dimension can randomly be mapped onto the cells of the cellular automaton (Figure \ref{fig:fig2}b). The size of the CA should follow the input feature dimension.

\begin{figure*}
\begin{center}
\includegraphics[width=0.7\textwidth]{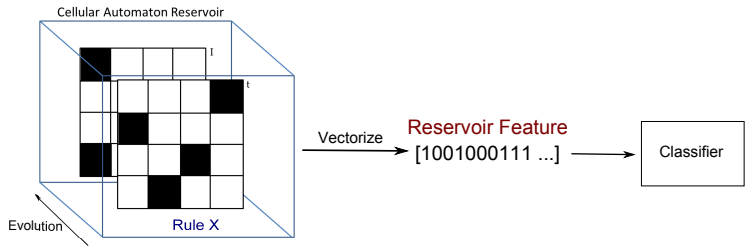} 
\end{center}
   \caption{Cellular automaton (CA) reservoir which is the space-time volume of the automaton evolution using Rule X. The whole evolution of the CA is vectorized and it is used as the reservoir feature for classification.}
\label{fig:fig3}
\end{figure*}

The cellular automaton evolves according to a prespecified rule (Figure \ref{fig:fig3}). It is experimentally observed that multiple random mappings are needed for accuracy. There are $R$ number of different random mappings, i.e. separate CA reservoirs, and they are combined into a large reservoir feature vector. The computation in CA takes place when cell activity due to nonzero initial values (i.e. input) mix and interact. Both prolonged evolution duration and existence of different random mappings increase the probability of long range interactions, hence improve computational power. 

\section{Results}
\label{results_section}
In order to test for long-short-term-memory capability of the proposed framework, 5 bit and 20 bit memory tasks were used. In these tasks, a sequence of binary vectors are presented, then following a distractor period, a cue signal is given after which the output should be the initially presented binary vectors. Input-output mapping is learned by estimating the linear regression weights via pseudo-inverse (see \cite{jaeger2012long} for details). These tasks have binary input, hence it is possible to randomly map the input values onto cells as initial states (Figure \ref{fig:fig2}b). Both 1D elementary CA rules and 2D Game of Life CA is explored. The total computational complexity is determined by the number of different random mappings, $R$, (i.e. separate reservoirs, Figure \ref{fig:fig3}) and the number of cell iterations $I$. The success criteria provided by \cite{jaeger2012long} is used in the experiments.   

\subsection{Game of Life}
5 bit task is run for distractor period $T_0$ 200 and 1000. The percent of trials that failed is shown for various $R$ and $I$ combinations in Table \ref{table:5BitGoL200and1000}. 20 bit task is run for distractor period $T_0$ 200 and 300, and again percent failed trials is shown in Table \ref{table:20BitGoL200and300}. It is observed that Game of Life CA is able to solve both 5 bit and 20 bit problems and for very long distractor periods. 
              
\begin{table}[h]                 
\centering                        
\begin{tabular}{l|l|l|l|l} 
$T_0 = 200$ & R=4 & 16 & 32 & 64 \\  
\hline     
I=4 & 100 & 100 & 100 & 4 \\
16 & 100 & 28 & 0 & 0 \\  
32 & 100 & 0 & 0 & 0 \\   
64 & 100 & 0 & 0 & 0 \\   
\end{tabular}                            
\quad                            						                                   
\begin{tabular}{l|l|l|l|l}   
 $T_0 = 1000$ & R=4 & 16 & 32 & 64 \\  
\hline      
I=4 & 100 & 100 & 100 & 100 \\
16 & 100 & 100 & 100 & 14 \\
32 & 100 & 100 & 100 & 0 \\ 
64 & 100 & 100 & 22 & 0 \\  
\end{tabular}                     
\caption{Percent failed trials for 5 Bit Task, Game of Life CA, $T_0 = 200$ (Left) $T_0 = 1000$ (Right). Rows are number of iterations $I$, and Columns are number of random permutations.}          
\label{table:5BitGoL200and1000}        
\end{table} 

\begin{table}[h]           
\centering                  
\begin{tabular}{l|l|l|l|l}      
 $T_0 = 200$ & R=192 & 256 & 320 & 384 \\ 
\hline  
I=12 & 100 & 100 & 100 & 32 \\
16 & 100 & 84 & 12 & 0 \\   
\end{tabular}                
\quad  
\begin{tabular}{l|l|l|l|l}      
 $T_0 = 300$ & R=192 & 256 & 320 & 384 \\ 
\hline  
I=12 & 100 & 100 & 100 & 60 \\
16 & 100 & 100 & 20 & 0 \\  
\end{tabular}               
\caption{Percent failed trials for 20 Bit Task, Game of Life CA, $T_0 = 200$ (Left) $T_0 = 300$ (Right). Rows and columns are the same as above.}    
\label{table:20BitGoL200and300}  
\end{table} 

\subsection{Elementary Cellular Automata}
5 bit task ($T_0 = 200$) is used to explore the capabilities of elementary cellular automata rules. Rules 32, 160, 4, 108,  218 and 250 are unable to give meaningful results for any [$R$, $I$] combination. Rules 22, 30, 126, 150, 182, 110, 54, 62, 90, 60 are able give 0 error for some combination. Best performances are observed for rules  90, 150, 182 and 22, in decreasing order (Table \ref{table:5BitElem}). It is again observed that computational power is enhanced with increasing either $R$ or $I$, thus multiplication of the two variables determine the overall performance. 

\begin{table}[h]           
\centering                  
\begin{tabular}{l|l|l|l|l}      
 $Rule 90$ & R=8 & 16 & 32 & 64 \\ 
\hline 
I=8 & 100 & 78 & 12 & 0 \\  
16 & 74 & 4 & 0 & 0 \\    
32 & 4 & 2 & 0 & - \\     
\end{tabular}                
\quad  
\begin{tabular}{l|l|l|l|l}      
 $Rule 150$ & R=8 & 16 & 32 & 64 \\ 
\hline 
I=8 & 100 & 80 & 8 & 0 \\   
16 & 84 & 6 & 0 & 0 \\    
32 & 8 & 0 & 0 & - \\
\end{tabular}                
\quad  
\begin{tabular}{l|l|l|l|l}      
 $Rule 182$ & R=8 & 16 & 32 & 64 \\ 
\hline 
I=8 & 100 & 82 & 18 & 0 \\  
16 & 92 & 14 & 0 & 0 \\   
32 & 12 & 0 & 0 & - \\  
\end{tabular}  
\quad
\begin{tabular}{l|l|l|l|l}      
 $Rule 22$ & R=8 & 16 & 32 & 64 \\ 
\hline 
I=8 & 100 & 78 & 20 & 0 \\  
16 & 86 & 16 & 0 & 0 \\   
32 & 16 & 0 & 0 & - \\   
\end{tabular} 
\caption{Percent failed trials for 5 Bit Task, Elementary CA Rules, $T_0 = 200$. Rows and columns are the same as above. }    
\label{table:5BitElem}  
\end{table} 

\begin{figure*}
\begin{center}
\includegraphics[width=0.5\textwidth]{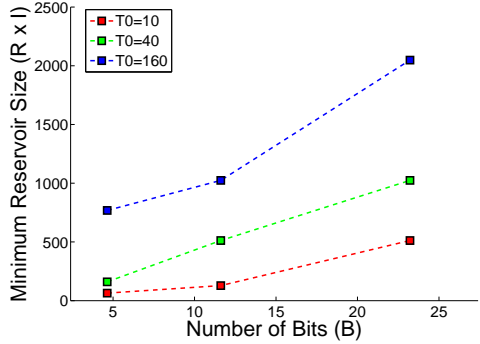} 
\end{center}
   \caption{Minimum resevoir size needed for zero error as a function of number of bits to be remembered. The experiment is repeated for 3 different distractor periods.}
\label{fig:fig4}
\end{figure*}

We further examined the performance of elementary cellular automaton rule 90 for $B$ bit memory  task ($B$ is a variable), using different reservoir sizes ($R \times I$) and distractor periods ($T_0$). The Figure \ref{fig:fig4} gives the minimum reservoir size to achieve zero error for different $B$ values. It is observed that there is a polynomial increase in the minimum required reservoir size with number of bits to be remembered and but a logarithmic increase with distractor period.   

In summary, the cellular automaton state space offers a rich and flexible reservoir of dynamical computation, that is capable of long-short-term-memory. Class 3 CA rules \cite{wolfram2002new}, which show random behavior, seem to give best performance in these type of tasks.

\subsection{Symbolic Processing with Additive Cellular Automaton Rules}
Some of the best performing CA rules explored above show additive behavior: the evolution for different initial conditions can be computed independently, then the results are combined by simply adding \cite{chaudhuri1997additive, wolfram2002new}. For example, rule 90 is additive under exclusive or (XOR) operation such that, when two separate initial conditions are combined by XOR (shown by $\oplus$), their subsequent evolution can also be combined by XOR. Although the combination logic for rule 150, and 22 (because it simulates rule 90) can also be derived, we will focus on rule 90.

Suppose we have two separate inputs, $A$, $B$. Let us assume that the nonzero entries in the input (i.e. initial states of the CA) represent the existence of categorical objects, as in 5 bit and 20 bit tasks. We compute the CA reservoir by applying the rule (i.e. 90) for a period of time steps and concatenating the state space, call them $C_A$ and $C_B$. We are interested in a new concept by combining the two inputs: $A \vee B$. This new concept should represent the union of objects, existing separately in $A$ and $B$, thus it is more abstract. Due to the binary categorical indicator nature of the input feature space, definition of logical combination rules are straightforward. We define OR operation by computing the reservoir of $A \vee B$:
\begin{align*}
    OR(A,B) = {C}_{A \vee B} = {C}_{A} \oplus {C}_{B} \oplus {C}_{A \wedge B} = {C}_{A} \oplus {C}_{B-A}.
\end{align*} 

The representation of the new concept obtained by union on existing concepts, can be computed \textbf{directly} on the cellular automata reservoir feature space via XOR operation, which is equivalent to addition operation on Galois Field, $F_2$. \footnote{($R$,$I$) parameter combination in our framework is analogous to aperture parameter in Conceptors.}

If the pattern is already stored in a concept, a repetitive addition does not make a change:
\begin{align*}
    {C}_{(A \vee B) \vee B} = {C}_{A \vee B} \oplus {C}_{\mathbf{0}} = {C}_{A \vee B},
\end{align*} 
and this is essential for incremental storage. 

Concept generated through NOT operation will arise from data which co-vary inversely:
\begin{align*}
    NOT(A) = {C}_{\neg A} = {C}_{A}.
\end{align*}

AND operation will generate a concept that consists of objects that co-exist in both $A$ and $B$ \footnote{Derived using De Morgan's rule and experimentally verified.}. :
\begin{align*}
    AND(A,B) = {C}_{A \wedge B} = {C}_{A} \oplus {C}_{A-B}.
\end{align*} 

XOR'ing and multiplication (in $F_2$) of existing concepts can be done: 
\begin{align*}
		XOR(A,B) = {C}_{A} \oplus {C}_{B},\\
    MULT(A,B) = {C}_{A} \wedge {C}_{B}.
\end{align*}

The classical rules of Boolean logic hold for this system (proofs not given). Overall the framework has great expressive power: availability of XOR and AND forms the whole $F_2$ field and it is possible to represent any logic obtainable by (OR , AND), with the additional benefit of algebraic operations. \footnote{Hence, Galois Linear Feedback Shift Registers (LFSR) are utilized.} However it should be noted that, the nice symbolic computation properties of the cellular automaton framework is applicable when the non-zero feature attributes of the CA initial states represent the existence of a predefined object/concept \footnote{Binary coding is another option but logical operations are meaningless in this case.}. Having said that, any input feature space can be transformed into this categorical indicator space by quantization or encoding using weighted summation (Figure \ref{fig:fig2}a), even though it might not be practical. Proper experiments should be designed to test the properties given above as in \cite{jaeger2014controlling}.   






\subsection{Computational Complexity}
There are two major savings of cellular automata framework compared to classical echo state networks:

1. Cellular automaton evolution is governed by bitwise operations instead of floating point multiplications. \\
2. Since the reservoir feature vector is binary, matrix multiplication needed in the linear classification/regression can be replaced by summation. 

Multiplication in echo state network is replaced with bitwise logic (eg. XOR for rule 90) and multiplication in classification/regression is replaced with summation. Overall, multiplication is completely avoided, both in reservoir computation and in classifier/regression stages, which makes the CA framework especially suitable for FPGA hardware implementations. 

\begin{table}[h]                 
\centering                        
\begin{tabular}{l|l|l|l} 
Task & ESN (Floating Point) & CA (Bitwise) & Speedup  \\  
\hline     
5 bit $T_0=200$ & 1.03 M & 0.43 M & 2.4X  \\
5 bit $T_0=1000$ & 13.1 M & 8.3 M  & 1.6X \\  
20 bit $T_0=200$ & 17.3 M & 9.5 M  & 1.8X \\   
\end{tabular}                     
\caption{The comparison of the number of operations for the echo state networks (ESN) and the Cellular Automata (CA) framework. Operation is floating point for ESN, but bitwise for CA.}
\label{table:ComparisonOperation}        
\end{table}

The number of operations needed for the reservoir computation of 5 bit and 20 bit tasks is given in Table \ref{table:ComparisonOperation}, both for Echo State Network (ESN) in \cite{jaeger2012long} and for cellular automata (CA). \footnote{For ESN, it is assumed that the number of floating point operations is equal to 2*NNZ.} There is a speedup in the number of operations in the order of 1.5-3X. However, considering the difference of complexity between floating point and bitwise operations, there is \textbf{almost two orders} of magnitude speedup/energy savings. \footnote{CPU architectures are optimized for arithmetic operations: bitwise logic takes 1 cycle and 32 bit floating point multiplication takes only 4 cycles, on 4th generation Haswell Intel\textsuperscript{TM} core. Therefore the speedup/energy savings due to the bitwise operations will be much more visible on hardware design, i.e. FPGA.}

For symbolic processing, there is \textbf{no additional computation} for CA framework, reservoir outputs can directly be combined using logical rules. However, Conceptors \cite{jaeger2014controlling} that are built upon ESN require correlation matrix computation and matrix multiplication of large matrices, for each input. As an example, for 20 bit task $T_0=200$, 1760 M floating point operations are needed for correlation matrix computation. Then there is a matrix inversion ($2000 \times 2000$ size, 68 M operations) and matrix multiplication (two $2000 \times 2000$ size matrices, 16000 M operations) to obtain the Conceptor matrix. All these computations (about 18 billion) are avoided by using an additive CA rule in our framework.

\section{Discussion}
\label{discussion_section}
We provide a novel framework of recurrent computation that is capable of long-short-term-memory and symbolic processing, which requires significantly less computation compared to echo state networks. \footnote{Comparison with other RNN algorithms are not provided but the computational complexity argument seems to be generally valid.} In the proposed reservoir computing approach, data are passed on a cellular automaton instead of an echo state network, and similar to echo state networks with sparse connections, the computation is local (only two neighbors in 1D) in the cellular automata space. Moreover, the best performing cellular automata rules are additive. How does extremely local, additive and bitwise computation gives surprisingly good performance in a pathological machine learning task? This question needs further examination, however the experiments suggest that if a dynamical system has universal computation capability, it can be utilized for difficult tasks that require recurrent processing once it is properly used. The trick that worked in the proposed framework is multiple random projections of the inputs that enhanced the probability of long range interactions. However it is useful, this expansion is expected to vastly increase the feature dimension for more complicated tasks, and curse with dimensionality.     

Cellular automata are very easy to implement in parallel hardware such as FPGA (\cite{halbach2004implementing}) or GPU (unpublished experiments on \cite{MarkFiserWeb}). 100 billion cell operations per second seem feasible on mid-range GPU cards, this is a very large number considering 10 million operations are needed for 20 bit task. Several theoretical advantages of the cellular automata framework compared to echo state networks are mentioned, in addition to their practical benefits. Cellular automata are easier to analyze, have insurances on Turing completeness and allows Boolean logic as well as algebra on Galois Field.  

There are a few extensions of the framework that is expected to improve the performance and computation time:\\
1. A hybrid \cite{sipper1998evolving} and a multilayer automaton can be used to handle different spatio-temporal scales in the input. \\
2. The best rule/random mapping combination can be searched in an unsupervised manner (pre-training). The rank of the binary state space can be used as the performance of combinations.\\
3. The problems due to the large dimensionality of the feature space can be alleviated by using a bagging approach that also selects a subset of the feature space in each bag \cite{latinne2000mixing}.\\
4. GPU programming can be devised to significantly (around 200X \cite{MarkFiserWeb}) speed up processing.

As a future work we would like to test the framework in real data tasks such as language modeling, music and handwriting prediction. Also symbolic processing performance of the cellular automata reservoir needs to be evaluated.


{\small
\bibliographystyle{IEEEtran}
\bibliography{egbib_oy}
}

%
%

\end{document}